\documentclass[letterpaper]{article} 
\usepackage{aaai24}  
\usepackage{times}  
\usepackage{helvet}  
\usepackage{courier}  
\usepackage[hyphens]{url}  
\usepackage{graphicx} 
\usepackage{natbib}  
\usepackage{caption} 
\usepackage{algorithm}
\usepackage{algorithmic}

\usepackage{newfloat}
\usepackage{listings}
\DeclareCaptionStyle{ruled}{labelfont=normalfont,labelsep=colon,strut=off} 
\floatstyle{ruled}
\newfloat{listing}{tb}{lst}{}
\floatname{listing}{Listing}
\newcommand{\sysname}{\textsc{\sffamily{Bait}}}
\newcommand\alg[1]{\textsc{\sffamily{#1}}}
\newcommand\system[1]{#1}
\newcommand\dataset[1]{#1}

\title{\sysname{}: Benchmarking (Embedding) Architectures for\\Interactive Theorem-Proving}

\author{
    Sean Lamont\textsuperscript{\rm 1, 2},
    Michael Norrish\textsuperscript{\rm 1},
    Amir Dezfouli\textsuperscript{\rm 3},
    Christian Walder\textsuperscript{\rm 4},
    Paul Montague\textsuperscript{\rm 2}
}
\affiliations{
    \textsuperscript{\rm 1}Australian National University\\
    \textsuperscript{\rm 2}Defence Science and Technology Group\\
    \textsuperscript{\rm 3}BIMLOGIQ\\
    \textsuperscript{\rm 4}Google DeepMind \\
    sean.lamont@anu.edu.au
}

\usepackage{amsfonts}
\usepackage{tabularx}

\usepackage{subcaption}
\usepackage{booktabs}
\usepackage{multirow}
\usepackage{amsmath}

\begin{document}

    \maketitle

    \begin{abstract}
        Artificial Intelligence for Theorem Proving has given rise to a plethora of benchmarks and
        methodologies, particularly in Interactive Theorem Proving (ITP).
        Research in the area is fragmented, with a diverse set of approaches being spread across several ITP systems.
        This presents a significant challenge to the comparison of methods, which are often complex and difficult to replicate.

        Addressing this, we present \sysname{},
        a framework for fair and streamlined
        comparison of learning approaches in ITP.
        We demonstrate \sysname{}'s capabilities with an in-depth comparison, across several ITP benchmarks, of state-of-the-art architectures applied to the problem of formula embedding.
        We find that Structure Aware Transformers perform particularly well, improving on techniques associated with the original problem sets.
        \sysname{} also allows us to assess the end-to-end proving performance of systems built on interactive environments.
        This unified perspective reveals a novel end-to-end system that improves on prior work.
        We also provide a qualitative analysis, illustrating that improved
        performance is associated with more semantically-aware embeddings.
        By streamlining the implementation and comparison of Machine Learning algorithms in the ITP context, we anticipate \sysname{} will be a springboard for future
        research.
    \end{abstract}

    \section{Introduction}
    Interactive Theorem Proving (ITP), a central paradigm of formal verification,
    has been used to write verified compilers~\cite{leroy_compcert_2014, tan_verified_2019},
    formalise mathematical conjectures~\cite{gonthier_four_2008}, and develop provably correct microkernels~\cite{klein_sel4_2009}.
    As proficiency in both the formal system and application domain is needed,
    large verification projects require significant human resources and expertise.
    This restricts their scalability and widespread adoption, with, for example, the seL4 verification taking 25 person years~\cite{klein_sel4_2009}.
    Advances in Artificial Intelligence for Interactive Theorem Proving (AI-ITP) have shown potential in automating
    and assisting human ITP guidance, but there remain many challenges in the area.
    With the current state of the art achieving $42\%$ accuracy on the (highly difficult) miniF2F-curriculum benchmark, there is still much progress to be made \cite{lample_hypertree_2022, zheng_minif2f_2021}.

    To that end, it is important to efficiently and fairly compare approaches.
    As mentioned in~\cite{yang_leandojo_2023},
    problems such as compute requirements and private code have made research in the area difficult.
    Adding to this difficulty is the fragmentation of results across ITP systems.
    Benchmarks and environments exist for \system{HOL Light} \cite{bansal_holist_2019, kaliszyk_holstep_2017}, \system{HOL4} \cite{wu_tacticzero_2021}, \system{Lean} \cite{polu_formal_2022, han_proof_2021, yang_leandojo_2023}, Isabelle \cite{li_isarstep_2020} and Metamath \cite{kaliszyk_mizar_2015}.
    These provide a broad set of tasks for benchmarking. However being isolated to a single system complicates comparisons between them.
    This is magnified by the variety and complexity of the learning algorithms, which vary over several axes.
    For example, \alg{TacticZero} \cite{wu_tacticzero_2021} uses a seq2seq autoencoder for expressions,
    and learns through online Reinforcement Learning (RL) with a custom goal selection algorithm.
    \cite{bansal_holist_2019} instead use Breadth First Search (BFS) for goal selection,
    with offline learning over labelled proof logs.

    \begin{table*}[ht]
        \centering
        \begin{tabular}{lccccccccc}
            & & \multicolumn{2}{c}{Tactics} & \multicolumn{3}{c}{Learning} & & \multicolumn{2}{c}{Representation} \\
            \cmidrule(r){3-4}\cmidrule(l){5-7}\cmidrule{9-10}

            Approach & ITP & Fixed & Gen
            & SL & RL & Pretrain & Search
            & Graph & Seq \\
            \toprule
            \citet{yang_learning_2019}                     & \multirow{2}{*}{\system{Coq}}  &            & \checkmark & \checkmark &  & & BestFS & \checkmark &\\
            \alg{GamePad} \cite{huang_gamepad_2018} & & & \checkmark & \checkmark & & & BestFS
            & \checkmark & \\
            \midrule
            \citet{bansal_learning_2019}$^{1}$             & \system{HOL Light}             & \checkmark &            & \checkmark &            &            & BFS & \checkmark &\\
            \midrule
            \alg{TacticZero} \cite{wu_tacticzero_2021}     & \multirow{2}{*}{\system{HOL4}} & \checkmark &  & & \checkmark & & Fringe &  & \checkmark\\
            \alg{TacticToe} \cite{gauthier_tactictoe_2021} &                                & \checkmark &            & \checkmark &            &            & BestFS  & & \checkmark \\
            \midrule
            \citet{wu_int_2020}                            & \system{INT}                   & \checkmark & \checkmark & \checkmark &            &            & MCTS    & \checkmark & \checkmark \\
            \midrule
            \citet{jiang_lisa_2021} & \system{Isabelle} & & \checkmark & \checkmark & & \checkmark
            & BestFS & & \checkmark \\
            \midrule
            \citet{polu_formal_2022}$^{2}$                 & \multirow{3}{*}{\system{Lean}} &            & \checkmark & \checkmark & & \checkmark & BestFS & & \checkmark\\
            \alg{ReProver} \cite{yang_leandojo_2023}       &                                & \checkmark & \checkmark & \checkmark &            & \checkmark & BestFS & & \checkmark\\
            \alg{HTPS} \cite{lample_hypertree_2022} & & & \checkmark & \checkmark & &
            \checkmark & MCTS* & & \checkmark \\
            \midrule
            \alg{Holophrasm} \cite{whalen_holophrasm_2016} & \system{Metamath}              & \checkmark & \checkmark & \checkmark & & & UCT* & & \checkmark \\
            \midrule
            \citet{poesia_peano_2023}                      & \system{Peano}                 & \checkmark &            &            & \checkmark &            & BestFS* &            & \checkmark \\
            \bottomrule
        \end{tabular}
        \caption*{$\sp{1}$Also \citet{paliwal_graph_2020,bansal_holist_2019}. $\sp{2}$Also \citet{polu_generative_2020, han_proof_2021}. *: Modified}
        \caption{An overview of current learning approaches in ITP.
        The ITP column indicates the underlying Proving System. The Tactics column indicates whether a generative model is used to generate the tactic. The Learning column indicates the use of Supervised Learning (SL), Reinforcement Learning (RL) and/or Pretrained Models. The Search column indicates the strategy for choosing which goal(s) to work on next. The Representation column indicates whether a graph or sequence representation is used. %
        The variety of approaches and underlying systems apparent here motivates our unified platform.}
        \label{benchmarks}
    \end{table*}%
    Despite this, we show that many fundamental components are common across AI-ITP systems.
    We leverage these to develop \sysname{}, the first cross-platform, unified framework for AI-ITP research.
    \sysname{} brings together several environments, datasets and models in AI-ITP, with a central interface
    for experiments and sharing components between systems.
    Being fully open source, the addition of new benchmarks and algorithms is facilitated
    by a modular and decoupled design.
    \sysname{} combines automatic logging, checkpointing and configuration management
    to allow for rapid testing and prototyping of ideas with minimal overhead.

    We use \sysname{} to study an important problem in AI-ITP, which is the choice of embedding architecture.
    The embedding model is critical, being used to encode ITP expressions
    for subsequent tactic, premise and goal selection.
    Current results either use graph based approaches \cite{kaliszyk_holstep_2017, paliwal_graph_2020, crouse_improving_2020}, or treat expressions as a sequence \cite{lample_hypertree_2022, polu_formal_2022, han_proof_2021},
    with no thorough comparison between them across ITP systems.
    INT~\cite{wu_int_2020} provides the only comparison, in a synthetic proving environment, without directly isolating the embedding architecture. We study approaches in various tasks across several ITPs, including recent Structure Aware Transformers~\cite{chen_structure-aware_2022, luo_transformers_2023}.

    \noindent To summarise, we present the following contributions:
    \begin{itemize}
        \item We introduce \sysname{}, the first cross-platform, open source
        framework for benchmarking results across ITP systems.
        \item We use \sysname{} to perform a comparison of embedding architectures in ITP, finding that Structure Aware Transformers \cite{chen_structure-aware_2022, luo_transformers_2023} improve upon Graph Neural Networks (GNNs) and Transformers across several datasets.
        \item We reveal a large improvement in the end-to-end \alg{TacticZero} \cite{wu_tacticzero_2021} algorithm through experiments with the embedding architecture, with a qualitative analysis finding more semantically-aware embeddings.
    \end{itemize}

    \section{Background}

    \subsection{Related Work}
    The recent \alg{LeanDojo}~\cite{yang_leandojo_2023} provides an open source ITP framework with easily reproducible results.
    As with other frameworks, such as \dataset{HOList}~\cite{bansal_holist_2019}, INT~\cite{wu_int_2020} and CoqGym~\cite{yang_learning_2019}, it is focused on a single proving system.
    Our contribution is complementary, building upon the extensive work developing these systems
    to create a unified cross-platform framework.
    MWPToolkit and LILA~\cite{lan_mwptoolkit_2022, mishra_lila_2022}
    unify approaches for Math Word Problems, a related area of AI for mathematical reasoning.

    The prevailing embedding models in AI-ITP are GNNs and Transformers, with the only direct comparison in the synthetic INT \cite{wu_int_2020} framework.
    We extend this to multiple benchmarks, and study combinations of both architectures.

    \subsection{Benchmarks and Approaches in AI-ITP}
    Current benchmarks are based either on proxy tasks from proof logs, or on end-to-end proving performance with ITP interaction.
    A common proxy task is to predict lemmas useful for a goal, which is an important step in many ITP tactics.
    Such \textit{premise selection} benchmarks include \dataset{HOLStep} \cite{kaliszyk_holstep_2017}, \dataset{MIZAR40} \cite{kaliszyk_mizar_2015} and ISARStep \cite{li_isarstep_2020}.
    \dataset{LeanStep}~\cite{han_proof_2021} 
    includes premise selection and other tasks such as predicting masked subterms or types.
    \dataset{HOList}~\cite{bansal_holist_2019} contains premise and tactic selection data in \system{HOL Light}, based on both human and synthetic proof logs, with an end-to-end environment for evaluation.
    Other end-to-end benchmarks include \dataset{CoqGym}~\cite{yang_learning_2019}, 
    \dataset{LeanGym}~\cite{polu_formal_2022},
    \dataset{LeanDojo}~\cite{yang_leandojo_2023}, 
    \dataset{LISA}~\cite{jiang_lisa_2021} for \system{Isabelle}, and \system{INT}~\cite{wu_int_2020}, a synthetic proving system for AI-ITP research.
    \dataset{miniF2F}~\cite{zheng_minif2f_2021} contains Olympiad, undergraduate and high school mathematics problems in \system{Lean}, \system{Metamath}, \system{HOL Light} and \system{Isabelle}. The difficulty of problems has made this a key benchmark for performance in state-of-the-art systems~\cite{lample_hypertree_2022, yang_leandojo_2023, polu_formal_2022}.

    From Table \ref{benchmarks}, we note the variety of learning approaches and their fragmentation across ITPs. For example, we see the exclusive use of fixed tactic models in \system{HOL4} and \system{HOL Light}, while generative models are prevalent in \system{Lean}, Isabelle and Coq.
    This divide in approaches motivates a cross-platform framework, with the particularly large split in graph and sequence based representations further motivating a study of embedding architectures.

    \section{AI-ITP}
    The ITP learning task can be modelled as a sequential decision problem~\cite{powell_reinforcement_2022, wu_tacticzero_2021}.
    The objective in each round is to prove an initial theorem, defined as the goal $g$.
    The initial state $(s_0 \supseteq g) \in \mathcal{S}$ includes $g$ along with auxiliary information such as allowed premises and axioms.
    The model $\pi: \mathcal{S} \to \mathcal{A}$ is a mapping from the state to an action $a \in \mathcal{A}$.
    The environment $\mathcal{E}: \mathcal{A} \to \mathcal{S}$ extends the ITP system,
    taking an action $a$ to a new state $s \in \mathcal{S}$, which
    is either a terminal state confirming the proof of the original goal,
    or represents a new state with subgoal(s) sufficient to prove $g$.
    We also assume that $s$ includes all previous states such that the model has a full view of the history, making the problem a Markov Decision Process (MDP). The full state therefore represents all currently observed paths to proving the original goal, referred to as the \textit{proof tree}.
    $\mathcal{E}$ can also include reward signals for reinforcement learning.

\begin{figure}[h]
        \centering
        \includegraphics[width=\linewidth]{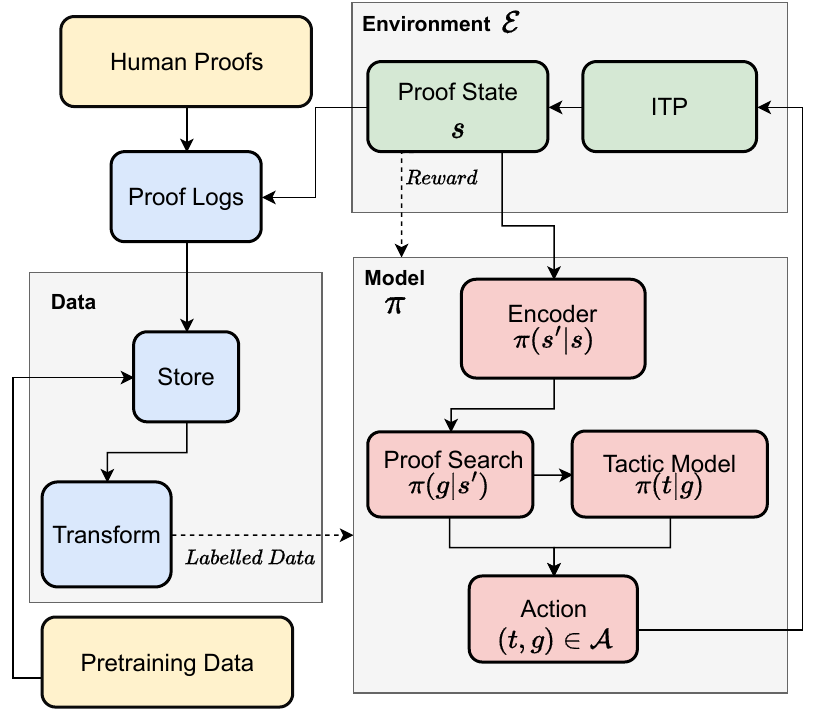}
        \caption{AI-ITP setup. A model $\pi$ interacts with a proving environment $\mathcal{E}$, mapping a state $s$ to an action $(t,g)$, which defines tactic(s) $t$ to apply to goal(s) $g \subseteq s$. $\pi$ is trained with rewards or Data from processed Proof Logs, sourced from human proofs or agent-environment interactions.}
        \label{fig:ai-itp}
    \end{figure}

    An AI-ITP system, such as the approaches in Table~\ref{benchmarks}, defines how the model $\pi$ is updated, given an interactive environment $\mathcal{E}$ and Data in the form of Proof Logs, as shown in Figure~\ref{fig:ai-itp}.

    \subsubsection{Learning Approach}
    The most common learning approach is supervised learning over proof logs, collected either from human proofs or from interaction of the AI system with the ITP environment. From these logs, labelled objectives such as premise and tactic selection are generated for training.
    Combining both approaches has been shown to be superior in several cases~\cite{lample_hypertree_2022, bansal_learning_2019, polu_formal_2022}.

    Reinforcement Learning (RL) has also been used, with the agent learning from a reward signal generated from interaction with the environment.
    Far less common than supervised approaches in ITP, RL has the advantage of removing biases in the labelling of data.
    For example, it is common to assign negative labels to premises not used in an original human proof,
    despite the potential for these to be useful in a different proof of the same conjecture~\cite{kaliszyk_holstep_2017, kaliszyk_mizar_2015, bansal_holist_2019}.
    However, RL can suffer from high variance in the gradient updates \cite{sutton_reinforcement_2018}, and requires appropriately shaping the reward to optimise proof performance.
    It remains unclear which is superior in AI-ITP, further motivating a centralised comparison platform.

    Pretrained language models have also been used, which are generally finetuned on proof data as in ~\cite{lample_hypertree_2022, polu_formal_2022,  han_proof_2021}.

    \subsubsection{Encoder \label{state_embed}}
    The encoder model $\pi(s' | s): \mathcal{S} \to \mathbb{R}^{D}$
    maps the current proof state $s$ into Euclidean space, with $D \in \mathbb{N}$ depending upon the approach.
    An effective embedding model is critical, as it is the only information used by the tactic and goal selection models.
    Approaches either treat the current goals $g \subseteq s$ as a Natural Language sequence \cite{lample_hypertree_2022, polu_formal_2022, wu_tacticzero_2021}, or use a graph based representation \cite{bansal_learning_2019, paliwal_graph_2020, wang_premise_2017, evans_can_2018}.
    The resulting embedding is $s' \in \mathbb{R}^{n \times k \times d}$ where $n, k, d$ are the number of
    goals, tokens and embedding dimensions respectively.
    The tokens are often pooled so as to produce a single vector for each goal, such that $s' \in \mathbb{R}^{n \times d}$.
    \begin{figure*}[htbp]
        \centering
        \includegraphics[width=0.7\textwidth]{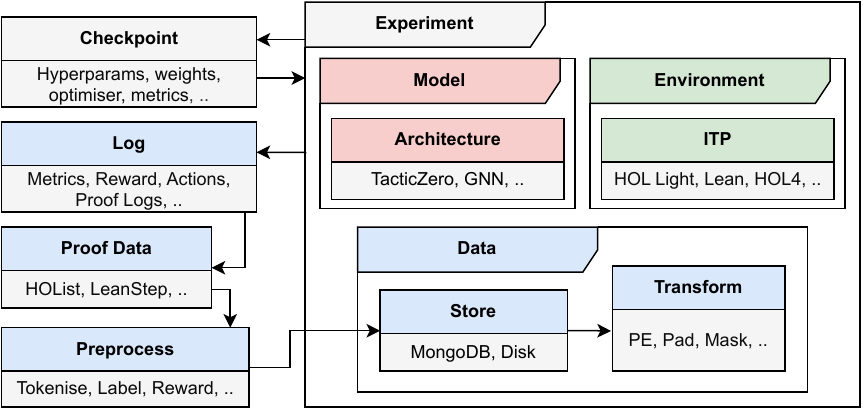}
        \caption{Overview of \sysname{}. The Experiment module abstractly defines tasks such as premise selection or RL.
        Experiment instances take a configuration specifying the Data, Model and Environment instances to use,
            with Checkpoints and Logs as output.
            Proof Data, either from existing sources or generated logs, is processed, stored and transformed for input to the Model.
        }
        \label{fig:framework}
    \end{figure*}

    \subsubsection{Proof Search}
    Proof search follows state embedding, as shown in Figure~\ref{fig:ai-itp}.
    It is the sub-model $\pi(g | s')$ mapping from the encoded state $s' \in \mathbb{R}^{ D}$
    to a subset of goals $g \subseteq s'$.
    A majority of approaches apply Breadth First Search \cite{bansal_holist_2019, huang_gamepad_2018} or Best First Search
    for goal selection.
    \alg{TacticZero} \cite{wu_tacticzero_2021} showed an improvement by considering the likelihood of proving distinct sets of goals (fringes)
    equivalent to the original goal.
    Improvements have also been found through variations of Monte Carlo Tree Search algorithms~\cite{lample_hypertree_2022, wu_int_2020}.

    \subsubsection{Tactic Model}
    Tactic selection maps the selected goals to an action, with $\pi(t | g) : \mathbb{R}^{D} \to \mathcal{T}$.
    ITP tactics are usually a small, fixed set, however
    they may include arguments which can be arbitrary expressions.
    It is therefore common to restrict arguments to only use other expressions, lemmas or terms available in the current context.
    Selecting these arguments is a particularly important problem, known as \emph{premise selection}.
    For these cases, the tactic model consists of predicting the tactic to apply,
    followed by the arguments conditioned on the tactic. Such approaches are marked as `Fixed' in Table \ref{benchmarks}.
    Despite a much larger output space, generative models such as Transformer Decoders have also
    been applied to predict the entire tactic and argument as a text sequence, listed under `Gen' in Table~\ref{benchmarks}. Holophrasm \cite{whalen_holophrasm_2016} and ReProver \cite{yang_leandojo_2023} combine these approaches, with premise arguments restricted to a fixed set and the remaining tactic tokens being generated.

    \section{\sysname{}}
    \sysname{}\footnote{\texttt{https://github.com/sean-lamont/bait}} is designed to be a general framework for AI-ITP,  with Data, Model and Environment modules implementing the setup in Figure~\ref{fig:ai-itp}. These are managed with an additional Experiment module, as outlined in Figure~\ref{fig:framework}.

    \subsubsection{Data}
    Data modules process, store and transform proof data.
    Raw data is sourced from human proofs and current datasets, or through proof logs generated by experiment runs.
    It is then pre-processed and stored. Pre-processing here includes, for example, generating labelled training splits, tokenisation and converting expressions into a standardised graph (or sequence) format.
    Final batch transformations, such as padding and masking for sequence models, are then applied by a data loader during an Experiment.

    \subsubsection{Environment}
    The Environment module embodies and extends the underlying ITP system. Support here includes converting model actions to the correct ITP format, processing the ITP state to be parsed by the model, and generating rewards for RL agents.

    \subsubsection{Model}
    This module implements architectures for the Encoder, Proof Search and Tactic Models in Figure \ref{fig:ai-itp}.
    Models are exposed to Experiments with a simple interface, which is straightforward to extend for integrating new architectures.

    \subsubsection{Experiment}
    Experiments in \sysname{} define tasks in AI-ITP. They take as input a Model, Data source and possibly Environment to interact with.
    We currently include Experiment classes for premise selection across several benchmarks, as well as the \dataset{HOList} \cite{paliwal_graph_2020} training and evaluation setup, \alg{TacticZero} RL setup \cite{wu_tacticzero_2021} and the INT \cite{wu_int_2020} experiments.

    Experiments inherit a single, shared implementation of logging and checkpointing to streamline their development.
    Logging is done using Weights and Biases \cite{wandb} which records all hyperparameters and metrics, with automatic plots, dashboards and run comparisons to ensure transparent and replicable results.

    Once implemented, Experiments are manged using Hydra \cite{yadan_hydra_2019}, which allows for complex configurations without modifying code. Hyperparameter sweeping is also simplified using these configurations.

    We provide more details and example usage of \sysname{} in the supplementary material.

    \section{Embedding Architectures}
    To demonstrate its utility as a research platform, we use \sysname{} to study embedding architectures in ITP.
    As we have noted, referring to Table~\ref{benchmarks}, there is no clear comparison of embedding approaches for different representations in ITP.
    We address this by comparing state-of-the-art approaches,
    for both graph and sequence representations,
    in several supervised and end-to-end ITP benchmarks.

    \subsection{Representations and Approaches}
    Embedding models in ITP are either graph or sequence based, depending on the underlying representation.

    \subsubsection{Sequences}
    Sequences in ITP are either human readable (pretty-printed), or structured s-expressions, which guarantee non-ambiguous operator precedence.

    \begin{table*}[ht]
        \centering
        \begin{tabular}[t]{lcccccc}
            \toprule

            Task                                 & SAT           & Directed SAT  & GNN  & Transformer   & Ensemble      & Bag of Words \\
            \midrule
            \dataset{HOList} Tactic              & 35.2          & 39.1          & 35.5 & 38.2          & \textbf{39.5} & 32.1         \\
            \dataset{HOList} Relative Parameter  & 98.0          & 98.2          & 98.3 & 98.3          & \textbf{98.7} & 98.0         \\
            \dataset{HOList} top-$5$             & 81.2          & \textbf{85.1} & 81.8 & 83.5          & 84.8          & 78.7         \\
            \dataset{MIZAR40} Premise Selection  & 73.6          & \textbf{74.1} & 73.5 & 73.7          & 73.5          & 72.3         \\
            \dataset{HOLStep} Premise Selection  & \textbf{90.9} & *             & 90.3 & 89.5          & 90.6          & 75.7         \\
            \dataset{LeanStep} Premise Selection & 96.6          & \textbf{97.0} & 95.7 & 95.9          & 96.1          & 92.1         \\
            \system{HOL4} Premise Selection      & 91.0          & 91.1          & 91.6 & \textbf{91.8} & 91.8          & 87.4         \\
            \bottomrule
        \end{tabular}
        \caption{Accuracy benchmarks for embedding architectures across Supervised Learning Tasks. $\sp{\mbox{*}}$Computationally intractable}
        \label{fig:results}
    \end{table*}%

    The current state-of-the-art model for sequences in ITP
    is the Transformer~\cite{vaswani_attention_2017}.
    The full Encoder-Decoder architecture has been used in end-to-end proving environments by~\cite{lample_hypertree_2022, wu_int_2020}.
    We also note that GPT style Decoder only architectures have been used in~\cite{polu_generative_2020, polu_formal_2022, han_proof_2021}, who predict all tactic tokens directly.
    As noted by~\cite{yang_leandojo_2023}, this is a bottleneck for premise selection and limits model generalisation, which was also observed in \cite{wu_int_2020}.
    Hence the Transformer Encoder remains an important part of the architecture, and is the model we use for our sequence experiments.

    \subsubsection{Graphs}
    For graph representations, expressions are parsed into abstract syntax trees, followed by various transformations to create a directed acyclic graph, such as variable renaming and subexpression sharing. See \cite{paliwal_graph_2020} or \cite{wang_premise_2017} for an explanation and comparison of different graph transformations, with additional details and examples in the supplementary material.

    GNNs are the current state-of-the-art for graph based forumula embeddings,
    improving upon LSTM, CNN and WaveNet based approaches across several benchmarks \cite{paliwal_graph_2020, wang_premise_2017, bansal_learning_2019, crouse_improving_2020}.
    The specific architectures used in ITP are generally variations on Message Passing Neural Networks (MPNN), adapted specifically for ITP formulae graphs~\cite{paliwal_graph_2020, wang_premise_2017, bansal_learning_2019}.
    We briefly summarise the basic concept here: \\
    Graph representations of expressions are given by $\mathcal{G} = (\mathcal{V}, \mathcal{E})$ with nodes $\mathcal{V}$ representing variables, constants and functions, and edges $\mathcal{E}$ mapping from functions to their arguments.
    The MPNN is parameterised by $T$ layers, or hops, with each layer aggregating \textit{messages} from the immediate neighbors in the graph.
    Messages are treated differently from node parents $\mathcal{N}^-$ and children $\mathcal{N}^+$, so as to model the directed nature of an expression graph.
    Edges $e_{ij} \in \mathbb{N}$ from node $i$ to node $j$ represent the order of arguments, and are initially projected into $\mathbb{R}^\mathbb{N}$ by a learnable embedding.
    At each step $t \in [0, T]$, the embedding $x_i$ of a node $i \in \mathcal{V}$ is
    \begin{align*}
        x_i^{t} = F_O\left(x_i^{t -1}, \mathcal{M}^t\right)
    \end{align*}
    With message $\mathcal{M}^t$ given by
    \begin{align*}
        F_A\left(x_i^{t-1}, \sum_{j\in \mathcal{N}^{-}(i)} F_P(x_j^{t-1}, e_{ji}), \sum_{j\in \mathcal{N}^{+}(i)} F_C(x_j^{t-1}, e_{ij})\right)
    \end{align*}
    $F_A, F_P, F_C, F_O$ are learnable functions, usually taken to be Multi Layer Perceptrons (MLPs), and the final node embeddings are the output of the final layer $\{x_i^T\}_{i\in\mathcal{V}}$.

    Structure Aware Transformer (\textit{SAT}) models~\cite{chen_structure-aware_2022} use the same graph representation as a GNN. The graph is first processed by an arbitrary GNN model, followed by a Transformer Encoder layer over the output. This is repeated for $T \in \mathbb{N}$ layers to produce the embeddings.
    \textit{Directed SAT} modifies this to restrict attention for each node to only include ancestors or descendent nodes, which has been found to improve performance for directed graph problems~\cite{luo_transformers_2023}.

    \section{Experiments}

    \subsubsection{Representations}
    For sequences, formulae are represented in s-expression format, which includes type information.

    For graphs, we take the s-expression sequences and parse them into abstract syntax trees. Graphs are then processed with subexpression sharing, as outlined in ~\cite{paliwal_graph_2020}.
    Using the same s-expression format in both sets of experiments ensures precedence, type and token information is kept constant between representations.

    \subsubsection{Model Details}
    Each of the $n\in\mathbb{N}$ tokens in an expression is initially represented with a one-hot vector,
    followed by an Embedding layer projecting these to the dimension $d\in \mathbb{N}$ of the network. The initial representations $\{x_i^0\}_{i=1}^{n}$ are then used as input for the embedding model. The final token embeddings $\{x\}_{i=1}^{n}$ are aggregated through sum, max or mean pooling to produce a single embedding vector $e \in \mathbb{R}^d$.

    Our Transformer Encoder follows the approach in~\cite{vaswani_attention_2017}, with sinusoidal positional encodings. We set a maximum sequence length of 1024.

    For GNNs, we use variations of the MPNN architecture in the previous section, as well as Graph Convolutional Networks (GCN)~\cite{zhang_graph_2019}.
    We also include results from a 0-layer GNN, as a simple Bag of Words baseline.

    Our SAT models use the implementations in \cite{chen_structure-aware_2022, luo_transformers_2023}, with no positional encoding.
    For Directed SAT, computing the ancestor/descendent nodes is required preprocessing, which can be expensive for large datasets.
    We were therefore unable to test this approach on the \dataset{HOLStep} benchmark.

    We also study an Ensemble which combines the best performing GNN and Transformer in each case. We implement this with an additional 2 layer MLP in the network, which concatenates the GNN and Transformer embeddings and projects them back into the original embedding dimension.

    \subsection{Supervised Setting}
    The benchmarks we use for supervised experiments are HOLStep~\cite{kaliszyk_holstep_2017}, MIZAR40~\cite{kaliszyk_mizar_2015}, LeanStep~\cite{han_proof_2021} and HOList~\cite{bansal_holist_2019}. We also include a new HOL4 premise selection task, which we generate using theorem dependencies from the HOL4 standard library.

    For premise selection experiments,
    embeddings are concatenated and fed through a 2-layer MLP
    with a final sigmoid layer and binary cross entropy as the loss.

    We perform a hyperparameter sweep over configurations for all architectures. The best scoring checkpoint on the validation set was then assessed on the test set.

    For \dataset{HOList} \cite{bansal_holist_2019}, we follow the exact supervised training setup described in \cite{paliwal_graph_2020}.
    A training point consists of a goal, tactic, positive premise and a list of negative premises, constructed by randomly sampling from the set of all used premises.
    The tasks are to predict the ground truth tactic and premise.
    We keep their loss function, which is a weighted sum over these objectives.
    The architecture, including the GNN, is identical to what is presented in~\cite{paliwal_graph_2020}.
    Table~\ref{fig:results} reports the tactic accuracy (Tactic), the number of times a positive premise is ranked higher than a negative (Relative Parameter), and whether the true tactic is in the top-5 predicted by the model (top-$5$) for the validation set.

    As in previous work, we use accuracy as the performance metric~\cite{paliwal_graph_2020, wang_premise_2017}.
    The full set of hyperparameters is in the supplementary material, with further details on the datasets and model architectures.

    \paragraph{Results}
    From Table~\ref{fig:results}, we see that the SAT models perform strongly in most benchmarks. The directed SAT model outperformed base SAT in all tested cases, highlighting the importance of modelling the directed structure. The \system{HOL4} task is an interesting exception.
    We hypothesise the small size of the dataset, with only 7000 unique expressions, is likely insufficient for the complex SAT model to learn beyond simpler approaches.
    Transformers outperformed GNN only approaches overall, and Ensembles almost always outperformed the base GNN and Transformer models on which they were based.
    The Bag of Words model performed poorly on all benchmarks with the exception of the \dataset{HOList} relative parameter accuracy.
    \cite{paliwal_graph_2020} also found good results using a Bag of Words model in \dataset{HOList}, when implemented as a 0 layer GNN.

    Despite the strong results,
    there are additional considerations to using SAT models in a practical AI-ITP system. It is far less efficient than current Transformer implementations, with a GNN pass in every layer complicating the optimisation of the attention computation.
    Directed SAT also requires computing ancestor and child nodes. Although this can be pre-computed, the vast majority of expressions generated in online agent-environment interaction are previously unseen.

    Our results are consistent with previous work, validating our implementation of \sysname{}.
    For \dataset{HOLStep}, the GNN we base on ~\cite{wang_premise_2017} performs similarly to their reported results.
    For \dataset{HOList}, our results are consistent with the numbers reported in ~\cite{bansal_holist_2019}, with `around 1\%' error in relative parameter accuracy (where we have 1.3\%--2\%), and a 39--41\% tactic accuracy. Their results also include synthetic proof logs to improve performance, while we only trained with human proof logs.
    The current state-of-the-art for the \dataset{HOLStep} and \dataset{MIZAR40} benchmarks remains the DAG-LSTM from~\cite{crouse_improving_2020}.
    As this architecture cannot compute separate embeddings for goals and premises, which makes prediction computationally intractable for end-to-end systems~\cite{wu_tacticzero_2021, paliwal_graph_2020, yang_leandojo_2023}, we omit this approach.

    \begin{table}[h]
        \centering
        \begin{tabular}{lcc}
            \toprule
            Model                     & Cumulative      & Valid (pass@1)  \\
            \midrule
            GNN                       & 96.2\%          & \textbf{64.6\%} \\
            Transformer               & \textbf{96.8\%} & 63.3\%          \\
            Original \alg{TacticZero} & 90.7\%          & 43.0\%          \\
            \bottomrule
        \end{tabular}
        \caption{Goals proven by \alg{TacticZero} in \system{HOL4}, after 1 attempt for validation and cumulatively for training.}
        \label{fig:tz_plotfirst}
    \end{table}%

    \begin{figure}[htbp]
        \centering
        \includegraphics[width=0.97\linewidth]{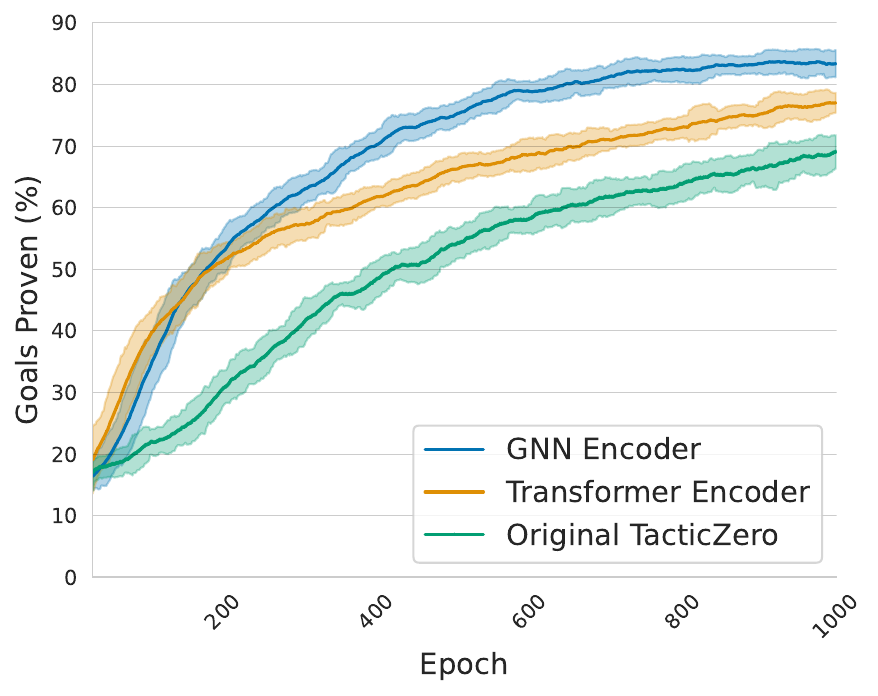}
        \caption{Goals proven during training by \alg{TacticZero}, with different underlying embedding architectures.}
        \label{fig:tz_plot}
    \end{figure}

    \renewcommand{\arraystretch}{1.5}
    \begin{table*}[ht]
        \centering
        \begin{tabular}[t]{lcr}
            \toprule
            Expression & GNN Encoder (Our Approach) & Original \alg{TacticZero} \\
            \midrule
            $\mathsf{diag}(A) = \mathsf{diag}(A^T)$ &
            $R = (R^T)^T$ &
            $\mathsf{FINITE}(\mathsf{POW}(s)) \Leftrightarrow \mathsf{FINITE}(s)$ \\
            $R\;x\; y \Rightarrow \mathsf{RC}(R)\; x\; y$ &
            $\mathsf{RC}(\mathsf{RC}(R)) = \mathsf{RC}(R)$ &
            $R \;x \;y \Rightarrow \mathsf{EQC}(R)\;x \;y$  \\
            $s \subseteq t \Leftrightarrow s \cup t = t$ &
            $ s\; \mathsf{DIFF}\; t = \emptyset \Leftrightarrow s \subseteq t$&
            $\mathsf{SURJ}\; f\; s\; t \Leftrightarrow \mathsf{IMAGE}\; f\; s = t$\\
            $ s \cup t = t \cup s$ &
            $ s \cup (t \cup u) = (s \cup t) \cup u$ &
            $s \cap t = t \cap s$ \\
            $ (s \cup t) x \Leftrightarrow x \in s
            \lor x \in t$ &
            $x \in s \cup t \;\Leftrightarrow\; x \in s \lor x \in t$ &
            $ (s \cap t) x \Leftrightarrow x \in s \land x \in t$ \\
            \bottomrule
        \end{tabular}
        \caption{A selection of mathematical expressions (left) along with the nearest expression by cosine distance according to the \alg{TacticZero} embedding (right) and GNN embedding (center).}
        \label{fig:qual1}
    \end{table*}%

    \subsection{End-to-End Setting}

    We use \alg{TacticZero} \cite{wu_tacticzero_2021} to study embedding architectures in an end-to-end setting.
    \alg{TacticZero} learns using only RL and interaction with the ITP environment.
    It is therefore a good benchmark for investigating embedding architectures beyond supervised problems.
    The original model uses a seq2seq autoencoder, trained on the reconstruction loss of the expressions.
    As the full training loop takes approximately 3 weeks, we test only a GNN and Transformer based Encoder.
    We pretrain these models on the \system{HOL4} premise selection task, ensuring goals used for this problem were excluded in the data generation.
    We take 1185 goals from the original paper, which are theorems from the \system{HOL4} standard library. We follow their protocol of a random 80:20 training and validation split, with 948 training goals and 237 validation goals.
    Figure~\ref{fig:tz_plot} plots the
    number of goals proven from the training set per epoch, and Table~\ref{fig:tz_plotfirst} tabulates the cumulative (proven at least once) and validation (pass@1) results, which are the standard metrics for end-to-end tasks~\cite{bansal_learning_2019}.

    \paragraph{Results}
    From Figure \ref{fig:tz_plot} and Table \ref{fig:tz_plot}, we observe a large improvement in \alg{TacticZero} when GNN or Transformer encoders are used, as compared to the original approach.
    The difference in validation performance is especially notable, with an approximate $50\%$ increase in goals proven by the GNN model compared to the original seq2seq autoencoder.
    This suggests that the autoencoder model is overfitting to the training set and struggling to generalise to unseen goals.

    As with our supervised experiments, when comparing the Transformer and GNN we find only a small difference in performance.
    INT~\cite{wu_int_2020}, found that Encoder-Decoder Transformers do not generalise as well as GNNs, which they suggest is due to the Decoder.
    For our case, we are only using a Transformer Encoder, with the rest of the architecture fixed.
    Our results are consistent with their hypothesis, with little difference between the GNN and Transformer (when excluding the Decoder) when generalising to unseen proofs in the Validation set.

    The embedding model used by the original \alg{TacticZero} is a
    fixed autoencoder trained only on the reconstruction loss of the original expression. Hence it is unsurprising to see an improvement in the end-to-end performance when compared to the other approaches. The original \alg{TacticZero} result reports $49.2\%$ of goals proven, compared to our observed $43\%$. Since we train for 200 additional epochs, it is possible that these additional epochs cause the model to overfit further. Regardless, our best approach is still a $31\%$ improvement over their result, which highlights the critical role of the embedding model
    in the overall system.

    \subsection{Qualitative Analysis}
    With the aim of explaining the large observed improvement in \alg{TacticZero}, we investigate the embeddings produced by the different models. We take a random set of expressions, generating their corresponding embeddings with the original \alg{TacticZero} autoencoder and the best performing GNN Encoder from Figure~\ref{fig:tz_plot}. We then find the nearest expression as judged by cosine distance in each of the embedding spaces.
    In a majority of cases, the nearest neighbor as judged by the GNN model was far more semantically relevant to the original expression than the nearest neighbor as judged by the original \alg{TacticZero} Autoencoder. Table \ref{fig:qual1} shows a small, insightful set of such cases, with a larger list in the Appendix (with full type information). We observe in several cases that the original \alg{TacticZero} encoder finds expressions which are not semantically similar at all,
    such as the third row in Table \ref{fig:qual1} where \alg{TacticZero}'s encoder mentions constants ($\mathsf{SURJ}$ and $\mathsf{IMAGE}$) that do not occur in the original expression, and fails to mention either subset or union, which do.
    In contrast, the GNN encoder finds an expression relating directly to subsets, and in this case it is easy to see how that expression might be useful in the proof of the original.
    We encourage the reader to compare other examples here and in the supplementary material.
    This provides some insight to the poor generalisation ability of the original encoder. If the nearest points in embedding space are semantically unrelated, then it is more likely for an unseen goal to be conflated with an unrelated expression.

    \subsection{Limitations}
    Due to computational constraints, we were unable to extensively evaluate HOList models on the end-to-end prover, with previous approaches using, for example, 2000 CPUs~\cite{bansal_learning_2019}. We nevertheless include a small scale evaluation in the supplementary material as a sanity check on our implementation.

    We did not evaluate several previous embedding approaches such as Tree-LSTMs or CNNs as there are several results indicating they are inferior to comparable GNN and Transformers~\cite{paliwal_graph_2020, wu_int_2020}.

    \section{Conclusion}
    We introduce \sysname{}, the first cross-platform framework for experimentation in Artificial Intelligence for Interactive Theorem Proving. Using \sysname{}, we compared modern embedding architectures over several supervised benchmarks, finding that Structure Aware Transformers and Ensemble approaches outperform GNN and Transformer baselines. Extending our analysis to end-to-end systems, we found a large improvement over previous work and observed more semantically accurate embeddings were produced as a result.

    \subsection{Future Work}
    For better coverage across different ITPs and learning algorithms, \sysname{} needs yet more benchmarks and datasets.
    \dataset{LeanDojo} \cite{yang_leandojo_2023} is a good candidate, being open source with state-of-the-art results on MiniF2F \cite{zheng_minif2f_2021}.
    \sysname{} was designed to compare all components of AI-ITP, so comparing the learning, tactic selection and proof search approaches in a similar manner is a natural next step.
    Given the growth in research of pretrained LLMs for ITP, integrating these is another promising direction.
    \sysname{} could also be used to investigate transfer tasks between systems,  such as~\cite{gauthier_sharing_2015}.

    \section{Acknowledgements}
    We would like to acknowledge Defence Science and Technology Group
    (DSTG) for their support in this project. We would also like to thank
    Minchao Wu for his help with the \alg{TacticZero} source code.

    \bibliography{ref}

\end{document}